\documentclass[letterpaper]{article} 
\usepackage{aaai2027}  
\usepackage[hyphens]{url}  
\usepackage{graphicx} 
\usepackage{natbib}  
\usepackage{caption} 
\usepackage{amsmath}
\usepackage{amssymb}
\usepackage{enumitem}
\usepackage{algorithm}
\usepackage{algorithmic}

\usepackage{newfloat}
\usepackage{listings}
\DeclareCaptionStyle{ruled}{labelfont=normalfont,labelsep=colon,strut=off} 
\floatstyle{ruled}
\newfloat{listing}{tb}{lst}{}
\floatname{listing}{Listing}

\usepackage{booktabs}
\usepackage{multirow}
\usepackage{subcaption}

\usepackage{tcolorbox}
\newcommand{\finding}{\textbf{Finding.}\ }
\newtcolorbox{findingbox}{
  colback=blue!6,
  colframe=blue!60!black,
  boxrule=0.5pt,
  leftrule=3pt,
  arc=2pt,
  left=6pt,
  right=6pt,
  top=5pt,
  bottom=5pt
}

\title{Reasoning Errors Have a Region and a Direction in the Residual-Stream Trajectory of LLMs}

\author{%
  \begin{tabular}{c}
    \textbf{Hamed Damirchi$^{1,2}$ \qquad Ignacio Meza De la Jara$^{1,2,3}$ \qquad Damith Ranasinghe$^{2,3}$} \\
    \textbf{Yuhang Liu$^{1,2,4}$ \qquad Javen Shi$^{1,2,4}$} \\[1ex]
    \textnormal{$^{1}$Australian Institute for Machine Learning \qquad $^{2}$Adelaide University \qquad
               $^{3}$Naval Group Pacific} \qquad \\ \textnormal{$^4$Responsible AI Research Centre, Australia} \\[0.5ex]
    \small{\texttt{\{firstname.lastname\}@adelaide.edu.au}}
  \end{tabular}
}
\affiliations{
}

\begin{document}

\maketitle

\begin{abstract}

As language models are increasingly used for tasks that require verifiable reasoning, reliably distinguishing sound reasoning from flawed reasoning has become an important practical problem. Recent trajectory-based methods seek this signal in layerwise residual-stream displacements, which capture how representations change while attenuating some stable, token-specific information. However, displacement omits the state from which an update originates, whereas restoring the full state risks reintroducing shortcut-prone information. We identify this trade-off and propose a three-stream detector that combines motion with two restricted views of location. A coarse region reader based on vector quantization and a fine direction reader over normalized multi-layer states. This design restores enough state context to interpret the motion without returning to full-state probing. On reasoning benchmarks unseen during training, our method improves selection accuracy by up to 12\% over the displacement-only state of the art and 21\% over single-layer probing baselines. Although trained only on reasoning benchmarks, it also reads factual completion and fact verification, ahead of every detector we compare against, which places the signal on correctness rather than on a kind of reasoning. Ablations further show that motion, region, and direction provide complementary signals. These results suggest that reasoning validity is better read from state-conditioned motion than from either static states or decontextualized trajectories alone.
\end{abstract}


\begin{figure}[t]
\centering
\includegraphics[width=0.95\columnwidth]{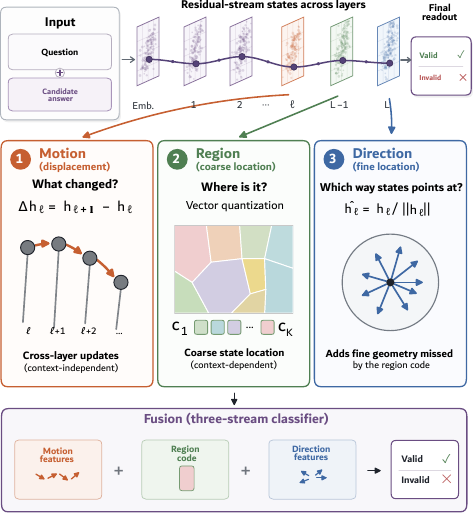}
\caption{\textbf{Three-stream validity detector.} Motion, quantized region, and direction signals are fused to classify candidate answers as valid or invalid.}
\label{fig:teaser}
\vspace{-1mm}
\end{figure}

\section{Introduction}
Language models are increasingly relied on for problems where reaching the correct answer requires reasoning, from a mathematical proof~\cite{ayoub2026structured} to the clinical reasoning behind a treatment recommendation~\cite{dekoninck2026openproofcorpuslargescale}. A user who cannot recheck every step is trusting that the model reasons soundly, and that it can tell a sound line of argument from a flawed one. That distinction is hard to audit from the answer alone. A wrong answer can read as fluently as a correct one, and the confidence a model assigns it does not track whether it is right~\cite{sun2025largelanguagemodelsoverconfident}. One place to look for a signal of reasoning soundness is in the internal representations the model forms as it reads it.

A common way to probe these representations is to train a linear classifier on activations from a single layer, treating each activation as a point and asking whether a linear boundary separates correct from incorrect answers~\cite{azizian2025the}. However, successful classification does not by itself establish that the probe has detected whether the underlying reasoning is sound. A single-layer representation may simultaneously contain two types of signal: 1) \textit{reasoning-related signals} that reflect the model's reasoning computation, and 2) \textit{label-correlated signals} that are predictive of the label but need not reflect reasoning validity, e.g., the wording, format, and topic of the input, token identity, and other dataset-specific regularities~\cite{bricken2023monosemantic, liu2025ipredictiam}. In this context, linear probing may exploit label-correlated signals because it is trained solely to minimize classification error. One empirical indication of this vulnerability is the poor cross-benchmark generalization of linear probes. A probe trained on one benchmark may fail when the wording, format, or topic changes~\cite{azizian2025the, postmus2024steering, rimsky-etal-2024-steering}.

Recent trajectory-based methods seek a more reliable readout by shifting attention from what information is present in a single-layer activation to how the representation changes across layers. Truth as a Trajectory~\cite{damirchi-etal-2026-truth}, for example, represents this evolution through layerwise displacements, the difference between the activation at one layer and that at the next. These displacements are unrolled across tokens and layers rather than reading the representation from a single layer. Intuitively, recent work suggests that signals remain recoverable from representations across layers~\cite{nikolaou2025language}, and the linear representation hypothesis further suggests that signals remain approximately linearly decodable across layers~\cite{gurnee2024language,turner2023steering}. Consequently, label-correlated signals that remain stable across adjacent layers may be suppressed by differencing, as their shared contributions cancel in the subtraction. Consistent with this intuition, trajectory-based methods have been shown to generalize better across benchmarks than single-layer probes~\cite{damirchi-etal-2026-truth}.

\begin{figure}[t]
\centering
\includegraphics[width=0.75\columnwidth]{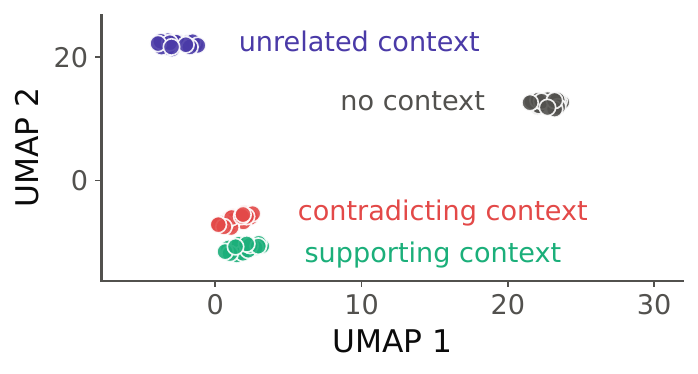}
\caption{Context changes the state of the same statement.
We vary only the context and observe that the resulting states occupy distinct regions of a UMAP projection, with supporting and contradicting contexts landing in different places.}
\label{fig:context}
\end{figure}

\paragraph{Challenges.}
While differencing in trajectory-based methods may suppress some label-correlated signals, it may also remove reasoning-related signals, particularly those encoding the model's reasoning state. Differencing suppresses information that remains stable across adjacent layers, and such stability need not separate label-correlated signals from reasoning-related ones. Such reasoning-state information is important for interpreting changes across layers, since the meaning of a representational change may depend on the state in which it occurs. As shown in Figure~\ref{fig:context}, supporting and contradicting contexts place otherwise matched statements in different regions of the representation space, indicating that layer states retain context-dependent validity information. This observation motivates restoring part of the state alongside the displacement. The key challenge is therefore to determine what state information, and how much of it, should be restored. Restoring too little may leave cross-layer changes difficult to interpret, whereas restoring the full state may reintroduce the label-correlated signals that differencing was intended to suppress. This creates a state-restoration trade-off, which we examine in Sec.~\ref{sec: tradeoff}.

\paragraph{Contributions.} To address the challenges above, we propose a framework that retains cross-layer changes while selectively restoring reasoning-related state information. It has three components. A motion reader follows trajectory-based methods and reads layerwise displacements, preserving their advantage in suppressing some label-correlated signals.

The other two are complementary state readers. The region reader projects normalized multi-layer states into a learned codebook and replaces each continuous state with its nearest codebook entry, retaining only a coarse state location. The direction reader instead preserves the normalized state directions across selected answer tokens and layers, retaining finer state distinctions while discarding magnitude and excluding the remaining tokens and layers. Together, the three readers balance the state-restoration trade-off by retaining cross-layer motion while restoring only the state information needed to interpret it. A reader designed to restrict state information need not remain restricted once trained, so we show that the restriction holds post-training. Our contributions are summarized as follows:
\begin{itemize}[leftmargin=*]
\item \textbf{We identify a state-restoration trade-off.} Trajectory-based differencing may discard reasoning-related state information, whereas restoring the full state may reintroduce label-correlated signals (Sec.~\ref{sec: tradeoff}).
\item \textbf{We address it with two complementary state readers.} A region reader restores coarse state location through codebook quantization, and a direction reader keeps finer state information as normalized per-layer directions (Sec.~\ref{sec: method}).
\item \textbf{What the readers restore transfers to benchmarks they never saw.} On reasoning benchmarks unseen during training, our method improves selection accuracy by up to 21\% over linear probing and 12\% over the displacement-only state of the art (Sec.~\ref{sec: exp}). Trained on reasoning alone, the same detector also outperforms every detector we compare against on unseen factual-completion and fact-verification benchmarks, which places what the streams read on correctness rather than on a kind of reasoning.
\end{itemize}

\section{Related Work}

Work on internal states commonly emphasizes one of two readings. One treats a hidden state as a static
point and asks whether correctness is encoded as a direction at some layer. The other treats the forward
pass as a process and reads how the state changes across layers. The first asks where the computation is;
the second asks how it moves. Our work joins these readings. We read the change, as the dynamic view does,
together with a compact reading of the location that differencing discards. We organize prior work around
these two readings and the point where ours brings them together.

\subsection{Static Linear Representations}
Linear representations are a central view in interpretability, where high-level properties appear as
directions in activation space \cite{park2023linear, elhage2022toy}. This view motivates tools with
different goals. Linear probes test whether a property is linearly decodable \cite{belinkov2022probing},
sparse autoencoders seek sparse, interpretable features \cite{cunningham2023sparse}, and
contrast-consistent search finds directions without labels \cite{burns2022discovering}. Applied
specifically to correctness, linear decoding yields a geometry of truth, a direction at a chosen layer
that separates correct from incorrect answers and can steer behavior when intervened on
\cite{marks2024geometry}. Two things limit it.
The direction is found by an unprincipled layer-by-layer search, and it is read from a static point that
mixes what the model computes with the surface form of the input. A probe can therefore rely on
dataset-specific vocabulary or format rather than a signal that transfers. Directions fitted for different
tasks can turn out close to orthogonal \cite{azizian2025the}, an effect attributed to the polysemantic
nature of activations \cite{lindsey2025biology}. We read the displacement trajectory rather than a static
point. Differencing attenuates stable surface information, and the compact location streams restore
context that the displacement drops. Together, they recover correctness structure that a single-layer
direction does not.

\subsection{Transformers as Trajectories}
A second view reads the forward pass as a dynamical system. The residual update $h_{\ell+1} = h_\ell +
f(h_\ell)$ motivates an Euler-integration view of depth \cite{chen2018neural, lu2019understanding}, so
the discrete layer-wise states can be read as a trajectory through activation space, and attention has been modeled as
interacting particles that cluster over depth \cite{geshkovski2023transformers}. Recent work models the
residual stream of a deployed model this way and reports that its effective rank collapses across depth
\cite{fernando2025transformerdynamics}. A related line reads depth as its own axis, decoding each layer's
state \cite{belrose2025elicitinglatentpredictionstransformers} or learning a recurrence over the stack of layers rather than over tokens
\cite{xu-etal-2024-rewiring}. The former decodes intermediate states, while the latter processes depth as a
sequence. Truth as a Trajectory \cite{damirchi-etal-2026-truth} follows the dynamical view for correctness,
reading the layer-to-layer displacement so the signal transfers across datasets. We keep that motion but
show that it under-determines the computation. The location streams instead read a set of selected depths
without a recurrence over depth.

\subsection{The Geometry of Reasoning}
A separate line of work also studies reasoning through the geometry of a trajectory and shares some of our
vocabulary, but it addresses a more structured setting. \citet{zhou2025geometry} split a passage of
reasoning into parts, average the final layer's activations across each part into one vector, and read the
sequence as a flow whose velocity tracks logical structure and whose position tracks the topic, a
demonstration carried out on deductions built for the study. Their flow keeps only the final layer of each
part and requires a passage already laid out as a chain of steps. This construction can describe the
geometry of a clean deduction, but it requires a problem-specific segmentation and evaluation setting.
Our detector instead reads supplied question-answer candidates as they occur in commonly used benchmarks,
with motion from their residual-stream displacements and compact location streams from their states. This
lets one detector train on a reasoning benchmark and select correct candidates on unseen reasoning
benchmarks and factuality tasks, without requiring an explicit chain of reasoning or a task-specific
segmentation.

\section{The Trade-Off in Restoring States}
\label{sec: tradeoff}
\begin{figure}[t]
\centering
\includegraphics[width=0.7\columnwidth]{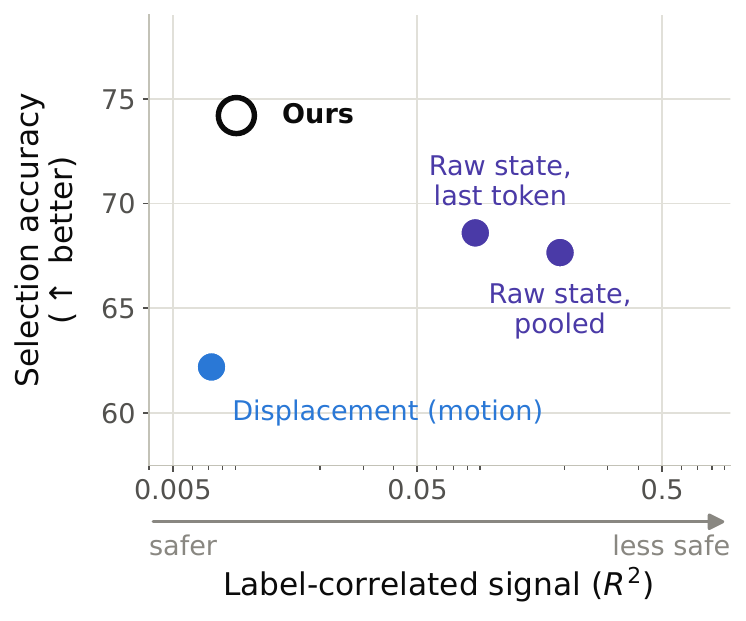}
\caption{\textbf{The trade-off the detector has to solve.} Each reader is placed by how often it
selects the correct answer on ARC-Challenge, and by how much of that answer's wording a held-out
regression recovers from what it hands its decision head. Ours is marked at whichever of its three
streams exposes the most wording. The two raw-state readers are measured at layer $14$, and accuracy
for the displacement reader and for ours is the in-distribution cell of Table~\ref{tab:main}.}
\label{fig:motivation}
\end{figure}

Differencing retains cross-layer changes but discards the state in which they occur. On ARC-Challenge
\cite{clark2018think}, a linear probe on one mid-network state selects the correct answer on $68.6\%$
of items, compared with $62.2\%$ for the displacement reader. The state therefore holds
correctness-relevant information that differencing does not pass on. But restoring the state also exposes
answer wording. At the probe's layer, a last-token state gives wording reconstruction $R^2{=}0.086$ and
retrieval $2.26$, compared with $0.007$ and $1.01$ for displacement; pooling every answer token raises
these to $0.191$ and $3.13$ while losing $0.9$ points of selection accuracy. The probe's donor advantage
also falls to $0.9$ points on held-out reasoning benchmarks (Table~\ref{tab:main}). Figure~\ref{fig:motivation}
shows the resulting trade-off. We therefore restore only the state information needed to interpret
cross-layer changes, rather than the full representation.

\section{A Three-Reader Framework}
\label{sec: method}
The previous section motivates a principle of \textit{minimally sufficient state restoration}, under which a detector retains cross-layer changes while exposing only the state information needed to interpret them. Figure~\ref{fig:teaser} shows the three-reader framework that instantiates it. A motion reader captures cross-layer changes through layerwise displacements, preserving the central advantage of trajectory-based methods. A region reader and a direction reader then selectively restore complementary forms of state information at different levels of detail. Their outputs are combined to produce a single validity score for each candidate answer.

\paragraph{Setup.} We work in the selection setting. A model is shown a question and a set of candidate answers, one correct and the rest not. For each candidate we pass the question and that candidate through the model and record its residual-stream activations, leaving the model unchanged. Write
$h_{t,\ell}\in\mathbb{R}^{d}$ for the activation at token $t$ and layer $\ell$, so that across the
$L$ layers of one pass these states form a trajectory through activation space. The detector scores
each candidate from that trajectory, and selects the highest-scoring one.

\subsection{Motion Reader for Cross-Layer Changes}
\label{sec: motion}
The motion reader is the model of prior work \cite{damirchi-etal-2026-truth}, which we take as it stands. It
reads the trajectory through its displacement,
\begin{equation}
d_{t,\ell} = h_{t,\ell+1} - h_{t,\ell} ,
\end{equation}
the update that layer $\ell$ writes into the residual stream. Differencing attenuates the persistent,
token-specific content the stream already holds, which is what lets the signal transfer across
datasets \cite{elhage2023privileged}. Stacking the displacements over every token and layer
gives one sequence, and a bidirectional LSTM reads it and returns its final state as the motion
embedding $c_{\mathrm{mot}}$. Nothing in this reading is new, and nothing about it is changed.

\subsection{Shared Representation for State Restoration}
The two location readers begin by giving up a state's size. The magnitude of the residual stream
grows with depth, so states drawn from different layers do not share a scale, and a reader that kept
it would pass that growth rather than the content. What they keep is the direction,
\begin{equation}
u_{t,\ell} = h_{t,\ell} / \lVert h_{t,\ell}\rVert ,
\end{equation}
and throughout, the direction of a state means this orientation, where the state itself points, not
the direction the trajectory travels in, which the motion already records. Both are confined to the
answer, since the question is common to every candidate, with $\mathcal{A}$ its tokens and $t^\star$
its final token. Both are also restricted to a fixed window $\mathcal{L}$ of $m = 6$ layers in the
middle-to-late range of the network. Early layers group tokens by surface form rather than by what
they mean \cite{nepal2025layer}, late layers work in the space of the token about to be emitted
\cite{yao2024circuits,dutta2024cot,xie2024calibrating}, and between the two, representations come
loose from the wording that produced them \cite{meng2022rome}, which is the surface detail the
previous section required a reading to leave behind. Our probe baseline agrees, settling between
layers $13$ and $16$ of $32$ for every donor when free to choose. The window is fixed in advance
rather than selected on results, and Table~\ref{tab:window} sweeps it. At an answer token, every
layer's direction passes through one projection $W$ shared across depth, and the results are
concatenated,
\begin{equation}
g_t = \big[\, W u_{t,\ell_1} \,;\, \dots \,;\, W u_{t,\ell_m} \,\big] .
\end{equation}
Normalization is what allows $W$ to be shared, since with magnitudes left in each slot would track a
layer's depth rather than its content, and depth survives because each layer keeps its own slot. The
two share this construction but not its weights, each fitting its own $W$, and differ only in how
much of $g_t$ each keeps.

\subsection{Region Reader for Coarse State Restoration}
\label{sec: region}
The region reader keeps only which part of the state space an answer occupies, not where inside that
part it falls. A learned codebook holds $K$ entries, trained jointly with the detector, and $K$ sets
the resolution. More entries pass a finer location, fewer a coarser one.

The reader uses the answer's final token, where the answer has settled. Its projected directions
$g_{t^\star}$ pass through an encoder into $z = \mathrm{Enc}(g_{t^\star})$, then the code is replaced by
the nearest entry $\{e_k\}_{k=1}^{K}$,
\begin{equation}
q(z) = e_{k^\star}, \qquad k^\star = \arg\min_{k}\, \lVert z - e_k \rVert .
\end{equation}
The head receives $q(z)$ as $c_{\mathrm{reg}}$, so two states assigned the same entry arrive as the
same input and no continuous code passes through the bottleneck. Quantization is hard for the head,
but learning uses the straight-through estimate
$\tilde{z} = z + \mathrm{sg}[\, e_{k^\star} - z \,]$ \cite{oord2017vqvae}. We initialize from data,
update entries by moving average, reseed unused entries, and use a commitment term to prevent codebook
collapse \cite{lancucki2020robust}; Appendix~\ref{app:regionusage} reports code usage.

\subsection{Direction Reader for Fine State Restoration}
\label{sec: direction}
A region names the part of space a state fell in and says nothing else, so a correct and an incorrect
answer landing in the same region reach the head as one answer. The direction reader keeps what that
discards, where the state sits inside the space rather than the name of the part it occupies.

It therefore keeps $g_t$ whole, recording where the state points at each depth in $\mathcal{L}$
separately rather than compressing the depths into one label. It has already given up the magnitude,
every layer outside $\mathcal{L}$ and every token outside the answer, so its narrowness comes from
what it declines to read.

A small MLP, shared across answer tokens, maps each $g_t$ to $p_t = \mathrm{MLP}(g_t)$. Nothing in
the concatenation relates one depth to another, and the MLP is what lets a pattern visible only
across depths be represented. An answer is a span rather than a point, so the reading combines two
summaries of it,
\begin{equation}
c_{\mathrm{dir}} = W_c \big[\, p_{t^\star} \,;\, \tfrac{1}{|\mathcal{A}|}\textstyle\sum_{t \in \mathcal{A}} p_t \,\big] .
\end{equation}
The final token holds the state the answer settles on, and the mean holds the states it passed
through on the way.

\subsection{Three Readers, One Decision}
Each reader returns one embedding. The three are concatenated and passed to an MLP head $f$ that
outputs a validity score,
\begin{equation}
\hat{y} = \sigma\big( f([\, c_{\mathrm{mot}} \,;\, c_{\mathrm{dir}} \,;\, c_{\mathrm{reg}} \,]) \big) ,
\end{equation}
and among a question's candidates the one with the highest $\hat{y}$ is selected. All parts train
together from scratch on the pairwise objective, with the commitment term added to the loss.

The embeddings from the readers are concatenated. Note that these embeddings are not symmetric in how
they treat depth, since the motion passes through the recurrence of an LSTM in layer order while the
two location readings treat the depths in $\mathcal{L}$ as an unordered set.

\section{Experimental Setup}
\label{sec: setup}
\paragraph{Models and features.}
All base-model weights remain frozen. We use Llama-3.1-8B \cite{llama3}, Qwen2.5-14B \cite{qwen25},
and Qwen3-30B-A3B \cite{qwen3}, spanning two families, $32$ and $48$ layers, and a mixture of experts
with $3$B active parameters. One forward pass per candidate records residual states at every layer and
token; every reader uses these activations. Appendix~\ref{app:models} gives the model depths and
extraction procedure.

\paragraph{Datasets.}
We use ARC-Challenge and ARC-Easy \cite{clark2018think}, OpenBookQA \cite{mihaylov2018suit},
CommonsenseQA \cite{talmor2019commonsenseqa}, Social IQa \cite{sap2019social}, and HellaSwag
\cite{zellers2019hellaswag} as donors, the one benchmark a detector trains on; evaluate reasoning
transfer on those six benchmarks plus Story
Cloze \cite{mostafazadeh2016corpus} and MMLU \cite{hendrycks2021measuring}; and test factual transfer
on FACTOR-wiki, FACTOR-news, FACTOR-expert \cite{muhlgay-etal-2024-generating}, and VitaminC
\cite{schuster-etal-2021-get}. TriviaQA and HaluEval are excluded.
Appendix~\ref{app:datasets} gives the dataset details.

\paragraph{Task and metric.}
Each item supplies one correct candidate and one or more incorrect ones. We train on one donor benchmark
and score each candidate from a frozen-model forward pass. Selection accuracy is the fraction of items
whose correct candidate outscores every incorrect candidate. We compare against a linear probe on one
validation-selected layer and the motion-only reader \cite{damirchi-etal-2026-truth}, using identical
items and splits. Appendix~\ref{app:protocol} gives the training recipe.

\section{Experiments}
\label{sec: exp}

Two questions come first. Trained on a single reasoning benchmark, does restoring the location improve
the detection of reasoning errors on benchmarks held out from training, and does the same detector
identify factual errors it was never shown? The rest ask where the gain comes from.

\subsection{Does restoring the location improve transfer to unseen reasoning benchmarks?}
We train a detector on one reasoning benchmark and evaluate it on the rest.
Table~\ref{tab:main} reports every training donor and method, one column per evaluation benchmark, with
the per-row average, the in-distribution score (ID), and the out-of-distribution average over the unseen
targets (OOD). Among the three trained readers the full model is best on $47$ of the $48$ cells, the
lone exception MMLU under the HellaSwag donor, where the motion-only reader edges ahead. Figure~\ref{fig:transfer-delta} in the
appendix maps that gain cell by cell, the difference between our full model and a motion-only reader.
The gain is spread across the grid rather than concentrated in a few donor and target pairs, and it is
largest on transfer into HellaSwag. Its out-of-distribution transfer
improves on the linear probe by $10$ to $21$ points across donors, and on the motion-only model of prior
work by $7$ to $12$, with its largest margin over prior work, $12$ points, on the ARC-Challenge donor.
Restoring the location is what moves the signal onto reasoning the detector was not trained on.
Across training seeds, the full model's per-donor reasoning means move by at most about a point, far
under the gap to either baseline (Table~\ref{tab:stability}).
Three controls place that gain on the design rather than on what came with it. Scaling the motion
reader to four times the width does not close it, removing magnitude from the motion reader does not
reproduce it, and the streams give back less of an answer's wording than the state a probe reads
(Appendix~\ref{app:design}).

\begin{table*}[t]
\centering \small
\setlength{\tabcolsep}{3pt}
\begin{tabular}{llccccccccccc}
\toprule
\multirow{2}{*}{\textbf{Train}} & \multirow{2}{*}{\textbf{Method}} & \multicolumn{8}{c}{\textbf{Evaluation benchmark}} & \multirow{2}{*}{\textbf{Avg}} & \multirow{2}{*}{\textbf{ID}} & \multirow{2}{*}{\textbf{OOD}} \\
\cmidrule(lr){3-10}
 & & ARC-C & ARC-E & OBQA & CSQA & SIQA & HSwag & MMLU & Story & & & \\
\midrule
     & Base Model & 45.4 & 57.6 & 50.6 & 44.7 & 44.5 & 79.0 & 40.8 & 77.9 & 55.1 & -- & -- \\
\midrule
    \multirow{3}{*}{ARC-C} & Linear Probe & 69.0 & 79.3 & 67.0 & 50.9 & 52.6 & 62.1 & 45.1 & 91.3 & 64.7 & 69.0 & 64.1 \\
    & Motion (TaT) & 62.2 & 73.4 & 62.6 & 61.5 & 53.4 & 61.3 & 47.0 & 83.1 & 63.1 & 62.2 & 63.2 \\
    & \textbf{Ours} & \textbf{74.2} & \textbf{86.3} & \textbf{79.1} & \textbf{66.7} & \textbf{61.2} & \textbf{81.4} & \textbf{55.0} & \textbf{96.9} & \textbf{75.1} & \textbf{74.2} & \textbf{75.2} \\
    \midrule
    \multirow{3}{*}{ARC-E} & Linear Probe & 66.2 & 84.6 & 72.6 & 56.3 & 51.4 & 64.3 & 48.5 & 88.3 & 66.5 & 84.6 & 64.0 \\
    & Motion (TaT) & 67.0 & 83.9 & 70.2 & 66.0 & 57.0 & 70.5 & 49.6 & 89.0 & 69.2 & 83.9 & 67.0 \\
    & \textbf{Ours} & \textbf{72.4} & \textbf{89.5} & \textbf{81.7} & \textbf{67.8} & \textbf{60.4} & \textbf{84.2} & \textbf{54.5} & \textbf{96.2} & \textbf{75.8} & \textbf{89.5} & \textbf{73.9} \\
    \midrule
    \multirow{3}{*}{OBQA} & Linear Probe & 57.8 & 64.6 & 79.8 & 44.7 & 49.2 & 52.2 & 39.6 & 83.6 & 58.9 & 79.8 & 56.0 \\
    & Motion (TaT) & 60.0 & 70.9 & 86.0 & 60.3 & 55.8 & 46.6 & 44.6 & 87.2 & 63.9 & 86.0 & 60.8 \\
    & \textbf{Ours} & \textbf{71.9} & \textbf{78.8} & \textbf{89.8} & \textbf{64.5} & \textbf{60.7} & \textbf{69.7} & \textbf{50.9} & \textbf{96.0} & \textbf{72.8} & \textbf{89.8} & \textbf{70.4} \\
    \midrule
    \multirow{3}{*}{CSQA} & Linear Probe & 50.5 & 66.2 & 66.6 & 72.5 & 52.6 & 47.1 & 38.9 & 84.9 & 59.9 & 72.5 & 58.1 \\
    & Motion (TaT) & 61.9 & 76.5 & 71.4 & 75.7 & 56.0 & 57.7 & 47.4 & 88.9 & 66.9 & 75.7 & 65.7 \\
    & \textbf{Ours} & \textbf{67.3} & \textbf{84.2} & \textbf{75.2} & \textbf{76.8} & \textbf{59.2} & \textbf{75.8} & \textbf{50.4} & \textbf{96.8} & \textbf{73.2} & \textbf{76.8} & \textbf{72.7} \\
    \midrule
    \multirow{3}{*}{SIQA} & Linear Probe & 53.6 & 64.6 & 66.6 & 47.2 & 64.5 & 31.6 & 40.5 & 83.0 & 56.5 & 64.5 & 55.3 \\
    & Motion (TaT) & 60.2 & 72.1 & 70.0 & 55.7 & 65.8 & 46.9 & 43.7 & 89.7 & 63.0 & 65.8 & 62.6 \\
    & \textbf{Ours} & \textbf{65.0} & \textbf{81.4} & \textbf{80.3} & \textbf{62.4} & \textbf{67.8} & \textbf{71.2} & \textbf{49.0} & \textbf{95.0} & \textbf{71.5} & \textbf{67.8} & \textbf{72.0} \\
    \midrule
    \multirow{3}{*}{HSwag} & Linear Probe & 40.3 & 45.3 & 41.6 & 38.2 & 41.6 & 86.4 & 27.1 & 85.4 & 50.7 & 86.4 & 45.6 \\
    & Motion (TaT) & 52.0 & 65.0 & 51.2 & 57.7 & 47.7 & 90.1 & \textbf{40.3} & 93.4 & 62.2 & 90.1 & 58.2 \\
    & \textbf{Ours} & \textbf{61.1} & \textbf{77.0} & \textbf{72.0} & \textbf{66.6} & \textbf{56.2} & \textbf{93.4} & 38.1 & \textbf{95.8} & \textbf{70.0} & \textbf{93.4} & \textbf{66.7} \\
\bottomrule
\end{tabular}
\caption{Cross-task transfer of reasoning-error detection, selection accuracy. Each detector trains on
the row benchmark and is scored on every column. \textbf{Avg} is the row mean over the eight targets,
\textbf{ID} the score on the training benchmark, and \textbf{OOD} the mean over the unseen ones, so
every off-diagonal cell is zero-shot transfer. Best per cell in bold, among the three trained methods.
Base Model is the frozen model's own answer likelihood, with no donor and no ID. On Llama-3.1-8B, Ours
is the three-seed mean (Table~\ref{tab:stability}).}
\label{tab:main}
\end{table*}

\begin{findingbox}
\finding What the location streams restore is not specific to the benchmark it was learned on. Every
donor yields a detector that reads reasoning it never trained on more accurately than both the
motion-only method and a probe.
\end{findingbox}

\subsection{Does what the detector reads extend beyond reasoning?}
Nothing in training touches factual claims, yet a reasoning-trained detector identifies factual errors it
was never shown. Table~\ref{tab:factual} puts reasoning-trained detectors on two factual benchmarks of
different form, with no further training. In FACTOR \cite{muhlgay-etal-2024-generating} the wrong option is a
minimal single-fact edit of a correct completion. In VitaminC \cite{schuster-etal-2021-get} a claim is
checked against a piece of evidence. We report the three donors that transfer best on reasoning, with the
grid for every donor in Table~\ref{tab:factualdonor}. The full model is the best of the three trained
readers on all four targets, ahead of the motion-only model by $9$ to $19$ points, and that lead holds in
every one of the twelve donor cells behind the means.
The frozen model's own likelihood sits beside them, and it reads the FACTOR sets above the full model
while trailing it by $18$ points on reasoning, since a single-fact edit of a fluent completion is
close to an unlikely one whereas a reasoning error is not (Appendix~\ref{app:likelihood}).

The two baselines trade places across the four targets. The probe reads FACTOR-expert and VitaminC better than the
displacement reader does, while the displacement reader leads on the two completion sets. The full model
is above both on every target, so its reach does not depend on which baseline suits a given set.
Averaging over all six donors rather than three narrows the gain but leaves it positive, since a donor
that transfers weakly on reasoning reaches the factual sets weakly too. Some benchmarks we considered were
read accurately by every method, ours included. We exclude TriviaQA and HaluEval because a shortest-answer
heuristic, without access to the model, reaches $88.8$ and $95.9$ selection accuracy; every reported
target is below chance under the same heuristic (Appendix~\ref{app:surface}). A linear read
trained on each factual set peaks at layers $12$ to $16$ (Figure~\ref{fig:midlayer}), inside the band
the location streams already read.
Neither result is particular to Llama-3.1-8B, and on a dense Qwen2.5-14B and a Qwen3-30B-A3B mixture
of experts the full model stays the best of the three readers on both axes, by $8$ to $14$ points on
reasoning and $11$ to $16$ on the factual sets (Appendix~\ref{app:crossmodel}).

\begin{table}[t]
\centering \small
\setlength{\tabcolsep}{4pt}
\begin{tabular}{lrrrr|r}
\toprule
Factual target & Probe & Motion & Ours & Length & Base Model \\
\midrule
FACTOR-wiki & 30.0 & 39.4 & \textbf{51.9} & 3.0 & 57.9 \\
FACTOR-news & 31.2 & 44.9 & \textbf{58.2} & 1.4 & 72.3 \\
FACTOR-expert & 55.5 & 50.8 & \textbf{69.5} & 5.1 & 74.6 \\
VitaminC & 62.3 & 57.6 & \textbf{67.1} & 23.2 & 65.1 \\
\bottomrule
\end{tabular}
\caption{The reach into factual errors, selection accuracy. Reasoning-trained detectors, scored
zero-shot on four factual targets (full test sets) and averaged over the three donors that transfer
best on reasoning (ARC-Challenge, ARC-Easy, CommonsenseQA). Length picks the shortest candidate, and
Base Model, set apart on the right, is the frozen model's own answer likelihood. Bold marks the best
trained detector. Single seed; per-donor cells in Table~\ref{tab:factualdonor}.}
\label{tab:factual}
\end{table}

\begin{findingbox}
\finding What the location streams read is closer to correctness than to any particular kind of
reasoning. A detector trained on reasoning alone, and never shown a factual claim, picks the true
completion out of minimal edits and checks a claim against its evidence.
\end{findingbox}

\subsection{Which added stream does the work?}
Table~\ref{tab:ablation} adds one location stream at a time to the motion base, on a single donor. Each
addition lifts the motion stream's performance on its own, the direction stream by $3$ to $20$
points across the eight reasoning targets and the region stream by $4$ to $15$ on seven of them,
leaving CommonsenseQA where it found it. The two together read best on every target. Removing the
direction stream from that pair costs $0.8$ to $5.7$ points and removing the region stream at most
$2.1$, though on HellaSwag the pair and the direction stream alone are level. \textbf{The gain rests on
neither stream alone.}
Whether the two recover the same answers is a question these accuracies do not settle.
A late-fusion variant that scores each stream on its own answers it, where every stream uniquely
recovers items the other two miss and a per-item oracle stands $3$ to $21$ points above the trained
combination (Appendix~\ref{app:latefusion}). Counting which streams read an item together shows the
region reading to sit almost inside the direction reading on reasoning but not on VitaminC, so which
readings agree is set by the kind of question asked.

\begin{table}[t]
\centering \small
\begin{tabular}{lrrrr}
\toprule
Target & Motion & $+$Direction & $+$Region & Ours \\
\midrule
ARC-Challenge & 62.2 & 73.0 & 73.8 & \textbf{74.7} \\
ARC-Easy & 73.4 & 85.3 & 83.0 & \textbf{86.5} \\
OpenBookQA & 62.6 & 79.2 & 76.8 & \textbf{79.4} \\
CommonsenseQA & 61.5 & 64.5 & 61.3 & \textbf{66.5} \\
Social IQa & 53.4 & 60.2 & 57.1 & \textbf{61.2} \\
HellaSwag & 61.3 & \textbf{81.7} & 76.0 & \textbf{81.7} \\
MMLU & 47.0 & 53.5 & 52.4 & \textbf{55.2} \\
Story Cloze & 83.1 & 96.5 & 96.1 & \textbf{96.9} \\
\bottomrule
\end{tabular}
\caption{Component ablation, selection accuracy, adding the direction stream, the region stream, or
both to the motion base, on the ARC-Challenge donor and the eight reasoning targets. All four columns
come from one seed, so Ours here is a single run rather than the three-seed mean of
Table~\ref{tab:main}. Best per row in bold.}
\label{tab:ablation}
\end{table}

\subsection{What does the region code hold?}
The region stream hands the head one of $128$ entries and nothing else. What a single entry is worth
depends on how much an answer's location says about whether it is correct.

\begin{figure}[t]
\centering
\includegraphics[width=\columnwidth]{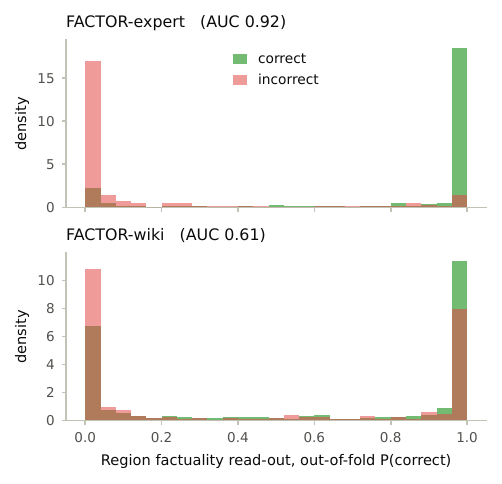}
\caption{A held-out linear read of the raw residual-stream state at layer $14$, out-of-fold, on two
FACTOR domains, against $0.5$ for a read that says nothing. The figure shows the state itself and not
the learned region code, so it measures the location signal available to be compressed rather than
what the region stream keeps. Layer $14$ is the depth the probe baseline most often selects.}
\label{fig:region-separation}
\vspace{-3mm}
\end{figure}

Before any compression, a linear classifier on the raw residual-stream state at layer $14$, scored on
held-out items, tells a correct completion from an incorrect one on FACTOR-expert at an AUC of $0.92$,
against $0.5$ for a read that says nothing. On FACTOR-wiki the same classifier reaches $0.61$
(Figure~\ref{fig:region-separation}), so how much location holds depends on the benchmark.

Part of that signal survives compression to a single entry. Predicting correctness from the region
alone beats the base rate on all four sets (Table~\ref{tab:region}). The same table counts the entries
each set uses, and Appendix~\ref{app:regionusage} maps which ones. No entry belongs to the factual sets
alone, and every entry a factual answer takes is one a reasoning answer takes as well. \textbf{The
detector was never given a vocabulary for factual answers and did not need one, which is part of why it
reads errors it was not trained on.}

\begin{table}[t]
\centering \small
\begin{tabular}{lrrr}
\toprule
Dataset & Codes used & Region$\,\to\,$correct & Base rate \\
\midrule
ARC-Challenge & 105 & 81.2 & 75.0 \\
Story Cloze & 42 & 88.2 & 50.0 \\
FACTOR-expert & 41 & 76.1 & 75.0 \\
VitaminC & 32 & 62.8 & 54.4 \\
\bottomrule
\end{tabular}
\caption{The region code as a shared vocabulary. Distinct codes used of the $128$, and the accuracy of
predicting correctness from the region alone (majority vote) against the base rate. One donor's
codebook (ARC-Challenge), single seed, up to $800$ items per dataset; the majority vote is a lower
bound on the code's signal.}
\label{tab:region}
\vspace{-3mm}
\end{table}

\subsection{Where in the network does the location have to be read?}
The middle third of the stack gives the highest reasoning mean, $75.6$, compared with $63.7$ early
and $69.5$ late, while all layers give the highest factual mean, $63.7$. Six layers in the middle
third also read higher than all $32$ on reasoning, $75.6$ against $74.0$, so which depths are read
matters more than how many. We retain the window fixed before these runs rather than selecting it on
the sweep. Appendix~\ref{app:window} gives the controlled comparison and its implications for the
reasoning and factual axes.

\section{Limitations}

Our detector is evaluated only by selecting among supplied candidates, not by judging whether a
single answer is correct. It needs the residual stream, a forward pass per candidate, a labeled donor
benchmark, and separate training for each model. Its performance is bounded by how much the model
state itself distinguishes correct from incorrect answers, and this bound varies across target sets.
The location streams retain direction but discard activation magnitude, which may contain correctness
signal. Finally, late fusion does not use all information in the three streams. The gap between the
trained combination and a per-item oracle indicates room for a learned gate that selects or weighs
streams for each item. Appendix~\ref{app:limitations} gives the detailed evidence and discussion.

\section{Conclusion}
A displacement records how a computation moves and drops where it sits. We restore that location
beside the motion at two resolutions, a discrete region and a per-layer direction, and read all three
at the answer. Kept narrow by design, the two added streams are $6.7\%$ of the motion reader's
parameters and give back little of an answer's wording. Trained on one reasoning benchmark and scored
on benchmarks held out from training, the detector improves on a linear probe by $10$ to $21$ points
of selection accuracy and on the displacement reader by $7$ to $12$. Trained on reasoning alone, it
identifies factual errors it was never shown, both minimal single-fact edits and claims checked
against their evidence. Both results hold on a larger dense model and on a mixture of experts. The
three readings are right about different items, and the region an answer occupies is a coordinate
that reasoning and factual errors share. Where a computation sits and how it moves are separable
readings of one trajectory, and correctness is legible in both.

\bibliography{aaai2027}


\clearpage
\newpage
\appendix
\setcounter{secnumdepth}{2}

\begin{center}
{\large\bfseries Supplementary Material for\\
Reasoning Errors Have a Region and a Direction in the Residual-Stream Trajectory of LLMs\par}
\end{center}

\section{Models and Feature Extraction}
\label{app:models}
Table~\ref{tab:models} lists the three models and the depths each detector reads. All three are base
models rather than instruction-tuned ones, and every weight stays frozen. The six depths sit at the same
fractions of network depth in every model, spanning the middle-to-late band Sec.~\ref{sec: method}
motivates, so a $32$-layer and a $48$-layer model are read at matching relative positions.

\begin{table}[h]
\centering \small
\setlength{\tabcolsep}{4pt}
\begin{tabular}{lrrl}
\toprule
Model & Layers & Width & Depths read \\
\midrule
Llama-3.1-8B & 32 & 4096 & 8, 12, 16, 20, 24, 28 \\
Qwen2.5-14B & 48 & 5120 & 12, 18, 24, 30, 36, 42 \\
Qwen3-30B-A3B & 48 & 2048 & 12, 18, 24, 30, 36, 42 \\
\bottomrule
\end{tabular}
\caption{The three models. Llama-3.1-8B and Qwen2.5-14B are dense; Qwen3-30B-A3B is a mixture of experts
with $30$B parameters of which $3$B are active per token. Width is the residual-stream dimension. The
depths are the layers at which the direction and region streams read the state.}
\label{tab:models}
\end{table}

For each question and each candidate answer we run one forward pass over the question followed by that
candidate, and record the residual-stream state at the output of every layer for every token, which gives
one array of shape $[\,n_{\mathrm{tokens}} \times n_{\mathrm{layers}},\, d\,]$ per candidate, stored in
half precision. Nothing is written back into the model, so the same activations serve every method we
compare.

The three streams read different parts of that array. The motion stream reads the displacements over the
whole token-and-layer grid, all tokens and all layers. The direction and region streams read only the
answer's tokens, at the depths in Table~\ref{tab:models}; where an answer runs longer than $64$ tokens,
$\mathcal{A}$ is its last $64$. Llama-3.1-8B is scored on the full test set of every target. On the two
Qwen models each target is scored on up to $2{,}000$ items, and the same items are used for every method.

\section{Datasets}
\label{app:datasets}
Our evaluation is built around generalization. Each detector trains on a single benchmark and is then
scored on eight reasoning benchmarks and four factual ones, on the same splits for every method. We
call that one benchmark the detector's donor, since it alone supplies training data and every other
benchmark in the suite is held out from it.

Six of the reasoning benchmarks serve as donors in turn, and they ask for different forms of reasoning. ARC-Challenge and ARC-Easy \cite{clark2018think} are grade-school science
questions, the challenge partition holding the ones that retrieval baselines fail. OpenBookQA
\cite{mihaylov2018suit} pairs a science fact with the everyday knowledge needed to apply it. CommonsenseQA
\cite{talmor2019commonsenseqa} asks about ordinary objects and situations, Social IQa \cite{sap2019social}
about people's motives and reactions, and HellaSwag \cite{zellers2019hellaswag} about which continuation
of a described event is plausible. Transfer between these is transfer between kinds of reasoning. We cap
each training set at $10{,}000$ items so that no benchmark supplies an order of magnitude more data than
another, which lets a difference between two donors be read as a property of the benchmark rather than of
its size.

The other two reasoning benchmarks are conventionally used for evaluation rather than training, and we use
them the same way. The Story Cloze test \cite{mostafazadeh2016corpus} asks which of two endings completes a
short everyday story, and MMLU \cite{hendrycks2021measuring} spans $57$ academic and professional subjects.
Neither resembles what a donor supplies. They extend the evaluation in two directions the training
benchmarks do not reach, narrative coherence and breadth of subject matter.

If what the location streams read is a signal about correctness rather than about reasoning tasks, it
should reach errors of another kind, so the suite ends with two factual paradigms that no detector
trains on. In FACTOR \cite{muhlgay-etal-2024-generating} every wrong option is a minimal edit of a correct
completion, changing one fact and leaving the rest standing, across three domains, wiki, news and expert.
VitaminC \cite{schuster-etal-2021-get} instead pairs a claim with a piece of evidence and asks whether the
evidence bears the claim out. The two put a factual error in front of the detector in different forms.
Two benchmarks common in this area, TriviaQA and HaluEval, are left out of the suite, since both can be
scored well above chance without consulting the model (Appendix~\ref{app:surface}).

\section{Task, Metric, Baselines and Training}
\label{app:protocol}
\paragraph{Task and metric.}
Each item in every dataset gives one correct answer and one or more incorrect ones. We form a positive
instance from each, the question with its correct answer, and one negative per incorrect answer. Our
metric is selection accuracy, the fraction of items whose correct answer outscores every incorrect
answer of that item, with chance at $1/(N{+}1)$ for an item with $N$ incorrect answers.

\paragraph{Baselines.}
We compare against two baselines on the same splits and metric. The first is a linear probe on a single
layer's activations, the standard static reader, with the layer selected by the donor's own held-out
validation selection accuracy, which lands between layers $13$ and $16$ for every donor
(Appendix~\ref{app:probe}). The second is the motion stream alone, the displacement LSTM of prior work
\cite{damirchi-etal-2026-truth}, which is our model with the two location streams removed. The gap
between that baseline and the full model is what restoring the location adds.

\paragraph{Implementation.}
The motion stream is a two-layer LSTM with $128$ hidden units. The direction stream
projects each layer's state to $8$ dimensions and the region stream to $64$, and the region codebook
holds $K{=}128$ entries. All parts train jointly from scratch with Adam at learning rate $10^{-3}$
under a cosine schedule, batch size $16$ with gradient accumulation of $4$. A tenth of each donor's
training data is held out for validation. We train for seven epochs and keep the checkpoint with the
best validation accuracy, and the same recipe is used on all three models. Each seed sets the weight
initialization and the data-order shuffle and leaves the data itself unchanged. The main table's
full-model rows are the mean over three training seeds (Table~\ref{tab:stability}), and each other
analysis states its seed and donor status in its caption.

\paragraph{Compute and software.}
Extraction and training run on NVIDIA RTX A6000 GPUs with $48$\,GB of memory, under Ubuntu, in
PyTorch $2.7$ with Transformers $4.52$ and h5py $3.13$. The activations of each dataset are written
once to HDF5 in half precision and read back by every method, so no base model is run twice.

\section{Surface Confounds and Excluded Benchmarks}
\label{app:surface}
Two benchmarks common in this area, TriviaQA and HaluEval, are absent from our suite. A benchmark
whose wrong answers are systematically longer than its right ones can be scored without consulting
the model, and Table~\ref{tab:length} reports a heuristic that selects the shortest candidate. On
TriviaQA the correct answer is shorter by $3.3$ tokens on average and on HaluEval by $11.6$, so the
rule reaches $88.8$ and $95.9$. A score on either would say more about how the benchmark was built
than about what a detector reads, so we set both aside. Every benchmark we report sits below chance
under the same rule. FACTOR and VitaminC are length-clean by construction, since a wrong option is a
minimal edit of the right one.

Length is not the only route. HaluEval's correct answers are human-written and its incorrect ones
model-generated, so its two classes differ in authorship as well as in correctness. A bag-of-words
classifier over the answer text alone, with no access to the model, separates them at an AUC of
$0.966$. Either measurement on its own would be enough to set the benchmark aside.

\begin{table}[t]
\centering \small
\begin{tabular}{lrr}
\toprule
Target & Chance & Shortest-answer rule \\
\midrule
ARC-Challenge & 25.0 & 9.8 \\
ARC-Easy & 25.0 & 10.2 \\
OpenBookQA & 25.0 & 3.8 \\
CommonsenseQA & 20.0 & 4.9 \\
Social IQa & 33.3 & 22.6 \\
HellaSwag & 25.0 & 23.6 \\
MMLU & 25.0 & 12.8 \\
Story Cloze & 50.0 & 39.3 \\
FACTOR-wiki & 25.0 & 3.0 \\
FACTOR-news & 25.0 & 1.4 \\
FACTOR-expert & 25.0 & 5.1 \\
VitaminC & 25.0 & 23.2 \\
\midrule
TriviaQA (excluded) & 50.0 & 88.8 \\
HaluEval (excluded) & 50.0 & 95.9 \\
\bottomrule
\end{tabular}
\caption{Selection accuracy $\times 100$ of a heuristic that selects the shortest candidate, with no
access to the model. Chance is $1/(N{+}1)$ averaged over items, so it tracks each item's candidate
count. Every reported target sits below chance; the two excluded benchmarks are solved by length
alone.}
\label{tab:length}
\end{table}

\section{The Linear Probe Baseline}
\label{app:probe}
The probe reads one layer, so the layer must be chosen fairly. We select it by the donor's own held-out
validation selection accuracy, sweeping all $32$ layers, which lands in a stable mid-network band for
every donor (Table~\ref{tab:probelayer}). Selecting the layer on a single factual set instead is noisy
and lands earlier in the network, at layers $9$ to $13$, because donor-to-factual selection accuracy is
low ($28$ to $46$) and its argmax is weakly identified. An earlier criterion that selected the layer
on TriviaQA and HaluEval is retired with those benchmarks. On the two Qwen models the same criterion
sweeps all $48$ layers of each.

\begin{table}[t]
\centering \small
\begin{tabular}{lrr}
\toprule
Donor & Selected layer & Reasoning mean \\
\midrule
ARC-Challenge & 14 & 64.7 \\
ARC-Easy & 14 & 66.5 \\
OpenBookQA & 13 & 58.9 \\
CommonsenseQA & 15 & 59.9 \\
Social IQa & 14 & 56.5 \\
HellaSwag & 16 & 50.7 \\
\bottomrule
\end{tabular}
\caption{The probe's validation-selected layer per donor and its cross-target reasoning mean, selection
accuracy $\times 100$. The choice is deterministic given the donor split.}
\label{tab:probelayer}
\end{table}

\section{The Model's Own Likelihood}
\label{app:likelihood}
The likelihood baseline reads no internal state. For each question and candidate we run one forward
pass over the frozen base model and score the candidate by the log-probability it assigns to the
answer's tokens alone, taking the highest scoring candidate as the pick. Nothing is trained, and the
prompts and the items are the ones every other row is scored on.

On reasoning the model's own likelihood is the weakest reader on average in Table~\ref{tab:main}. It
averages $55.1$ over the eight targets, where the probe averages $59.5$ across donors, the
displacement reader $64.7$, and the full model $73.1$, which puts it $18$ points below the full
model. \textbf{What the detector reads here is therefore a signal from the representations that shows
how the base model can distinguish the validity of candidates despite token space not allowing for
this signal to come through.}

The factual sets go the other way. The likelihood reads all three FACTOR sets above the full model,
by $6.0$, $14.1$ and $5.1$ points (Table~\ref{tab:factual}), and it does so under every scoring
variant we ran. FACTOR's wrong options are single-fact edits of an otherwise fluent completion, so
whether a completion is factual and whether it is likely nearly coincide, and a likelihood scorer is
reading the property the benchmark is built from. The one reasoning target it reads well is
HellaSwag, at $79.0$, which is a completion benchmark of the same shape. VitaminC asks something
else, whether a claim holds given a supplied passage, and there the full model leads, $67.1$ against
$65.1$. That margin is $2.0$ points on one set, and it holds on two of the three donors.
\textbf{Reasoning correctness is not as legible in the model's token space (what the model says) as
it is in how it computes, while the factual correctness of a minimal edit largely is.} Note that our
readers are trained on reasoning, and the assertion here is based on generalization to factual
benchmarks from reasoning datasets.

Three variants of that score come off the same forward pass, the total log-probability of the answer,
its per-token mean, and a per-character normalization. We report the per-token mean. A total inherits
the answer-length shortcut this suite is built to exclude, the same shortcut that sets TriviaQA and
HaluEval aside in Appendix~\ref{app:surface}, so headlining it would hold the baseline to a looser
standard than the paper holds its own targets. The ordering reported on the factual sets holds under
all three. This baseline is measured on Llama-3.1-8B, so the cross-model tables carry no such row.

\section{Per-Cell Transfer Gain}
\label{app:transfergain}
\begin{figure}[t]
\centering
\includegraphics[width=\columnwidth]{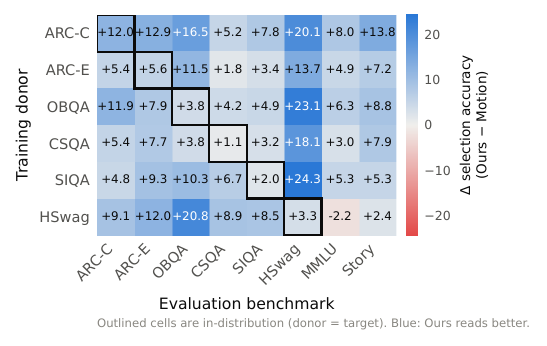}
\caption{Per-cell gain from restoring the location, Ours minus the motion-only reader on the transfer
grid of Table~\ref{tab:main} (selection accuracy $\times 100$). Blue is a gain for the full model, red a
loss. Outlined cells are in-distribution (donor equals target). The gain is broad and largest on transfer
into HellaSwag, with a single negative cell, MMLU under the HellaSwag donor.}
\label{fig:transfer-delta}
\end{figure}

Figure~\ref{fig:transfer-delta} reads Table~\ref{tab:main} as a map of the gain from restoring the
location, the full model minus the motion-only reader in every cell of the transfer grid. The gain is
positive on all but one of the $48$ cells and largest on transfer into HellaSwag.

\section{Result Stability Across Seeds}
\label{app:seeds}
Across three training seeds (weight initialization and data-order shuffle, identical data), the full
model's results move little. Table~\ref{tab:stability} reports the cross-target reasoning mean and
its spread for each training donor. The spread is a tenth of a point on the steadiest donors and about a
point at its widest, on HellaSwag and ARC-Easy, far below the roughly eight-point margin the full model
holds over the motion-only baseline and the larger margin over the probe. The ordering over the baselines
is therefore not seed-fragile.

\begin{table}[t]
\centering \small
\begin{tabular}{lcc}
\toprule
Donor & Reasoning mean $\pm$ std & Range \\
\midrule
ARC-Challenge & $75.1 \pm 0.2$ & $74.9$--$75.3$ \\
ARC-Easy & $75.8 \pm 0.9$ & $75.0$--$76.7$ \\
OpenBookQA & $72.8 \pm 0.1$ & $72.7$--$72.9$ \\
CommonsenseQA & $73.2 \pm 0.1$ & $73.1$--$73.3$ \\
Social IQa & $71.5 \pm 0.1$ & $71.4$--$71.7$ \\
HellaSwag & $70.0 \pm 1.0$ & $69.1$--$71.0$ \\
\midrule
Cross-donor mean & $73.1 \pm 0.4$ & $72.7$--$73.5$ \\
\bottomrule
\end{tabular}
\caption{Seed stability of the full model on the reasoning axis, selection accuracy $\times 100$. For
each training donor, the cross-target reasoning mean across three training seeds (weight initialization
and data-order shuffle, identical data), as mean $\pm$ standard deviation with the observed range. The
probe is deterministic given the split.}
\label{tab:stability}
\end{table}

\section{Factual Transfer per Donor}
\label{app:factualdonor}
Table~\ref{tab:factualdonor} breaks the factual means into donor cells. The full model beats the
motion-only model in all twelve cells, by $5.7$ to $30.1$ points, and beats the probe in eleven of
twelve, the exception FACTOR-expert under the ARC-Easy donor, where the probe reads $0.4$ higher on a
$236$-item set. The reach into factual errors requires a donor that itself transfers well on
reasoning, and a donor that transfers weakly on reasoning reaches these sets weakly too.

\begin{table}[t]
\centering \small
\setlength{\tabcolsep}{4pt}
\begin{tabular}{llrrr}
\toprule
Target & Donor & Probe & Motion & Ours \\
\midrule
\multirow{3}{*}{FACTOR-wiki} & ARC-C & 35.9 & 38.1 & \textbf{54.2} \\
 & ARC-E & 25.5 & 42.8 & \textbf{51.8} \\
 & CSQA & 28.4 & 37.4 & \textbf{49.9} \\
\midrule
\multirow{3}{*}{FACTOR-news} & ARC-C & 41.9 & 44.5 & \textbf{59.4} \\
 & ARC-E & 26.1 & 45.6 & \textbf{56.2} \\
 & CSQA & 25.6 & 44.6 & \textbf{59.1} \\
\midrule
\multirow{3}{*}{FACTOR-expert} & ARC-C & 44.9 & 34.7 & \textbf{64.8} \\
 & ARC-E & \textbf{73.3} & 61.9 & 72.9 \\
 & CSQA & 48.3 & 55.9 & \textbf{70.8} \\
\midrule
\multirow{3}{*}{VitaminC} & ARC-C & 64.5 & 54.5 & \textbf{70.0} \\
 & ARC-E & 62.0 & 58.9 & \textbf{64.6} \\
 & CSQA & 60.4 & 59.5 & \textbf{66.8} \\
\bottomrule
\end{tabular}
\caption{Per-donor factual transfer, selection accuracy $\times 100$, single seed, full test sets. The
FACTOR-expert rows rest on $236$ items, so their per-donor cells are noisy; read the means in the main
table.}
\label{tab:factualdonor}
\end{table}

\begin{figure}[t]
\centering
\includegraphics[width=\columnwidth]{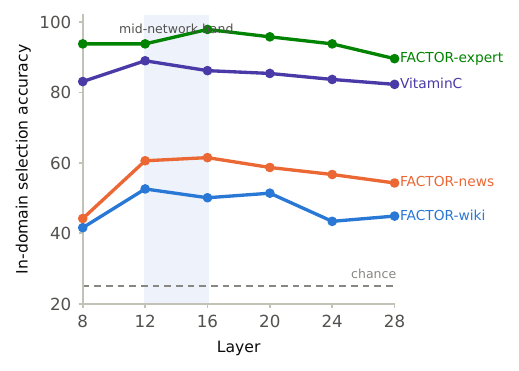}
\caption{In-domain per-layer selection accuracy of a linear read on each factual set, sweeping the six
extracted depths. Every set peaks in the mid-network band, layers $12$ to $16$. The FACTOR-expert curve
rests on about $47$ held-out items and is optimistic. Chance is $25$.}
\label{fig:midlayer}
\end{figure}

An in-domain linear read of each factual set, sweeping the six extracted layers, peaks at layer $12$
on FACTOR-wiki ($52.6$), layer $16$ on FACTOR-news ($61.5$) and FACTOR-expert ($97.9$), and layer $12$
on VitaminC ($89.0$). The FACTOR-expert ceiling rests on a held-out split of about $47$ items and is
optimistic. The sweep covers only the six extracted depths, so it locates which of those carries the
most in-domain signal rather than showing the band itself to be the best available
(Figure~\ref{fig:midlayer}).

\section{The Late-Fusion Decomposition}
\label{app:latefusion}
The detector of Sec.~\ref{sec: method} joins the three readings by concatenating their embeddings and
passing them to a single head, which returns one score and no per-stream quantity. Reading a stream on
its own therefore needs a model built to expose one. We train a variant that leaves every stream
unchanged and replaces the joint head with one linear head per stream. Writing $c_s$ for the embedding
of stream $s \in \{\mathrm{mot},\mathrm{dir},\mathrm{reg}\}$, each head returns a scalar,
\begin{equation}
o_s = w_s^{\top} c_s + b_s .
\end{equation}
Each logit is standardized on its own,
\begin{equation}
\tilde{o}_s = \frac{o_s - \mu_s}{\sqrt{v_s + \epsilon}} ,
\end{equation}
where $\mu_s$ and $v_s$ are the mean and variance of that stream's logit, taken over the batch during
training and from running estimates at evaluation. Nothing is learned in this step, neither a scale
nor a shift. The three standardized logits are then combined by a learned weight,
\begin{equation}
\hat{y} = \sigma\Big( \textstyle\sum_{s} \alpha_s \tilde{o}_s \Big), \qquad
\alpha = \mathrm{softmax}(a), \quad a \in \mathbb{R}^{3},
\end{equation}
with $a$ initialized so that the three weights start equal. All parts train jointly from scratch on
the pairwise objective, under the recipe of Appendix~\ref{app:protocol}.

The standardization is what makes the per-stream reading worth taking. Left out, one head can settle
at a near-constant logit while the others grow, and that stream then scores near the floor for reasons
of scale rather than of signal. What it does to a stream is a single affine map with a positive scale,
so it cannot reorder the candidates of an item. It changes what the heads learn, not how a stream
ranks the answers in front of it.

\begin{figure}[t]
\centering
\includegraphics[width=\columnwidth]{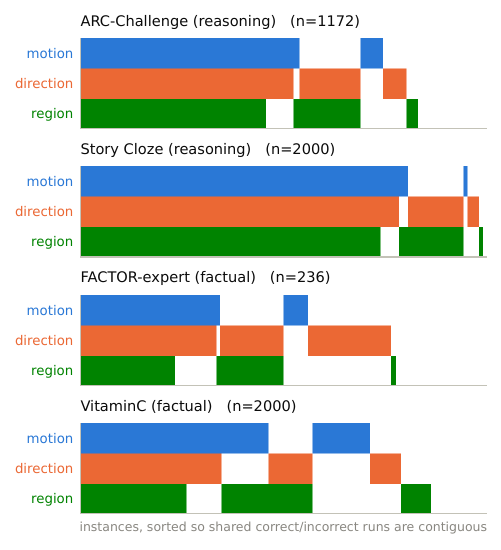}
\caption{The same per-stream correctness sets, item by item, colored by stream and sorted so shared
correct-and-incorrect patterns run together. Each column is one item, and a stream's row is filled
where that stream reads the item correctly on its own. One donor (ARC-Challenge), single seed, up to
$2{,}000$ items per dataset.}
\label{fig:stream-raster}
\end{figure}

Three quantities are read from the trained variant. A stream reads an item correctly when its own
logit on the correct candidate exceeds its logit on every incorrect candidate of that item, which
gives the per-stream columns of Table~\ref{tab:complement} and the correctness sets behind
Figures~\ref{fig:stream-overlap} and~\ref{fig:stream-raster}. The fused column is the model's own
combined score. The oracle counts an item correct when at least one of the three streams reads it
correctly, which bounds what a rule choosing between the streams item by item could reach; it is not
attainable without already knowing the answer.

The variant is a measurement device and not a second proposal. On the ARC-Challenge donor's own test
set it reaches $75.1$ against $74.7$ for the concatenation model of Sec.~\ref{sec: method} at the same
seed, so the streams are separated inside a detector of the same strength. Its solo numbers describe
what each stream can support on its own, not a component of the concatenation model's score.

\begin{figure}[t]
\centering
\includegraphics[width=\columnwidth]{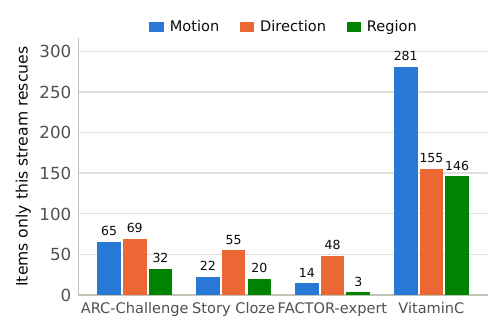}
\caption{Division of labor, the count of items each stream alone reads correctly when the other two miss
it, under the late-fusion decomposition of one trained detector. One donor (ARC-Challenge), single
seed, up to $2{,}000$ items per dataset.}
\label{fig:unique-saves}
\end{figure}

\begin{figure*}[t]
\centering
\includegraphics[width=1.99\columnwidth]{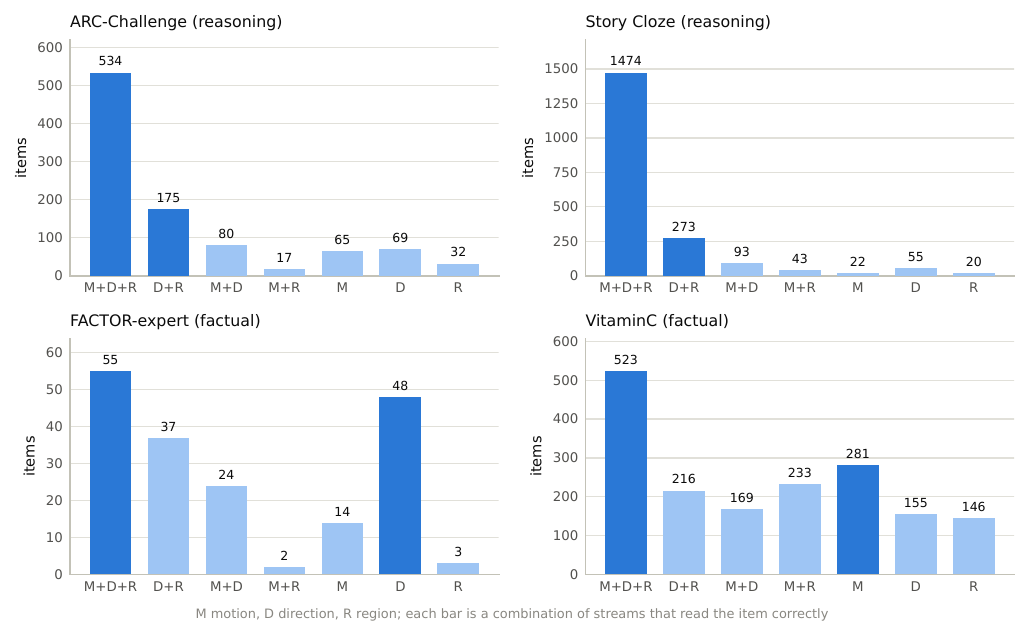}
\caption{Overlap of the items each stream reads correctly, under the late-fusion decomposition of one
trained detector. Each bar counts the items that a given combination of streams all read right, so the
single-stream bars reproduce Figure~\ref{fig:unique-saves}. The two largest groups of each panel are
shaded dark. M motion, D direction, R region. One donor (ARC-Challenge), single seed, up to
$2{,}000$ items per dataset.}
\label{fig:stream-overlap}
\end{figure*}

\paragraph{Are the streams redundant or complementary?}
The results in Table~\ref{tab:complement} show the direction stream to be the strongest single reader
on the two reasoning sets and on FACTOR-expert, while on VitaminC motion reads better than either
location stream. We also report an oracle over the three, which counts an item correct when any one of
them selects the right candidate and so bounds what choosing between the streams item by item could
reach. It clears the best single stream on all four sets, so signal sits in the three that no one of
them recovers alone. Figure~\ref{fig:unique-saves} shows where that signal is, with every stream,
motion included, uniquely recovering items the other two miss, and motion alone rescuing $281$ items
on VitaminC that neither location stream reads correctly.
Concatenation already turns part of that into accuracy, passing the best single stream on three of the
four sets. On FACTOR-expert it does not, reading $12$ points below the direction stream alone.
\textbf{Each reading is right about different items, so the three together hold more about correctness
than any reader here recovers from them, and the $3$ to $21$ points between the trained fusion and the
oracle measure how much.} Closing that gap asks for more than another rule for combining the three,
since gating and per-step injection score within $0.005$ AUC of concatenation
(Appendix~\ref{app:design}). The useful quantity is which of them to trust on a given item.

\begin{table}[t]
\centering \small
\setlength{\tabcolsep}{4pt}
\begin{tabular}{lrrrrr}
\toprule
Dataset & Motion & Direction & Region & Fused & Oracle \\
\midrule
ARC-Challenge & 59.4 & \textbf{73.2} & 64.7 & 75.1 & 82.9 \\
Story Cloze & 81.6 & \textbf{94.8} & 90.5 & 96.1 & 99.0 \\
FACTOR-expert & 40.3 & \textbf{69.5} & 41.1 & 57.2 & 77.5 \\
VitaminC & \textbf{60.3} & 53.1 & 55.9 & 65.0 & 86.2 \\
\bottomrule
\end{tabular}
\caption{Division of labor, selection accuracy. Each stream read on its own, the trained fusion of the
three, and a per-item oracle over the three, under the late-fusion decomposition of this appendix.
Bold marks the best single stream. One donor (ARC-Challenge), single seed, up to $2{,}000$ items per
dataset.}
\label{tab:complement}
\end{table}

\paragraph{Where do the streams' correct answers overlap?}
Each stream recovers items the other two miss. How the three group on the rest is a separate question,
and Figure~\ref{fig:stream-overlap} answers it by counting the items every combination of streams
reads correctly. On every dataset the largest group is the one all three read, so the streams are not
reading disjoint properties but one property with different reach. Beyond that core, the two reasoning
sets and FACTOR-expert are alike. What the region reading recovers there sits almost entirely inside
what direction recovers, since it seldom reads an item alone and seldom reads one alongside motion
that direction misses, so the coarse reading adds little the fine one does not already hold. VitaminC
does not behave that way. Region reads items on its own several times more often than on any of the
other three sets, and the items it reads together with motion while direction misses them are the
largest pair there. Figure~\ref{fig:stream-raster} shows the same sets item by item, where the wide
band of agreement on the reasoning benchmarks gives way to a divided one. \textbf{The coarse reading
is not a blurred copy of the fine one, and which readings agree is set by the kind of question asked
rather than by a fixed order among the three.} The component ablation of Table~\ref{tab:ablation} is
scored on reasoning alone, which is where the two location readings overlap most, so it reads the
region stream where it has least of its own to add.

\section{Design Checks and Controls}
\label{app:design}
\paragraph{Fusion.} On an earlier two-stream form of the model, concatenation, per-item gating and
per-step injection land within $0.005$ held-out AUC of one another on the reasoning targets we retain,
once checkpoints are selected by validation AUC (ARC-Challenge donor $0.732$, $0.736$, $0.734$), while
a unified attention head trails at $0.675$. We keep concatenation. It puts nothing learned between the
readings and the score, so what a reading contributes can be measured without also accounting for a
gate whose own behaviour varies from item to item.
\paragraph{Region codebook health.} A naive codebook collapses onto a handful of entries, and the
region stream then stops contributing. Seeding the entries from the first batch of data, updating them
by an exponential moving average, reseeding unused entries, and a commitment term holding the encoder to
its choice hold the codebook's perplexity near $50$ of the $128$ entries, where a fully collapsed
codebook would read $1$, and turn the region stream from a small loss into a gain on the factual axis.
The region stream is therefore worth having only if its codebook stays healthy.
\paragraph{Direction only.} The residual-stream norm grows with depth, so raw states from different
layers do not share a scale, and the location streams therefore read unit-normalized directions. Class
differences in the final-layer norm exist but are dataset-dependent, and reading the unnormalized state
in place of the direction scored lower in our development runs (see Limitations).

\paragraph{Is the gain just added capacity?}\label{app:capacity}
The two location streams add $0.31$M parameters to the motion reader, $6.7\%$ of its LSTM
(Appendix~\ref{app:params}). Small as that is, the gain might come from those parameters rather than
from location. We train both readers on the ARC-Challenge donor at hidden widths $128$,
$256$ and $512$ (Table~\ref{tab:size}). The full model leads at every width, by $9$ to $11$ points on
the reasoning mean and $13$ to $15$ on the factual one. Width does little for either. The full model
moves by $0.3$ points over a fourfold increase, while the motion-only reader gains $1.7$ points at
$256$ and gives most of them back at $512$. At that width its $25.7$M parameters read ten points below
the full model's $5.1$M. \textbf{The gap is the design, not the size.}

\begin{table}[t]
\centering \small
\begin{tabular}{llrrr}
\toprule
Hidden & Method & Params & Reasoning & Factual \\
\midrule
\multirow{2}{*}{$128$} & Motion & $4.8$M & $64.9$ & $46.7$ \\
 & \textbf{Ours} & $5.1$M & \textbf{75.6} & \textbf{61.8} \\
\midrule
\multirow{2}{*}{$256$} & Motion & $10.6$M & $66.6$ & $49.1$ \\
 & \textbf{Ours} & $11.3$M & \textbf{75.4} & \textbf{62.3} \\
\midrule
\multirow{2}{*}{$512$} & Motion & $25.7$M & $65.4$ & $48.1$ \\
 & \textbf{Ours} & $27.3$M & \textbf{75.3} & \textbf{61.5} \\
\bottomrule
\end{tabular}
\caption{Scaling both readers to hidden widths $128$, $256$, and $512$ on the ARC-Challenge donor,
selection accuracy. Reasoning is the mean over four reasoning sets (ARC-Challenge, ARC-Easy,
OpenBookQA, CommonsenseQA), Factual over FACTOR and VitaminC. Width scales each reader's recurrent
state, and the full model's stream widths with it; the codebook holds $128$ entries at every width.
Single seed, up to $2{,}000$ items per target.}
\label{tab:size}
\end{table}

\paragraph{Is the gain just normalization?}\label{app:normalization}
The location streams read direction-only states, so the gain might come from dropping magnitude rather
than from restoring location. We take the magnitude away from the motion reader as well, retraining it
on unit-length displacements (Table~\ref{tab:normmotion}). On the ARC-Challenge donor that costs $11$
points on the reasoning mean and leaves ARC-Easy where it was. On the factual mean, the normalized
reader loses $10$ points on the first donor and gains $2.5$ on the second, and the full model is
ahead of both motion readers on each. \textbf{The gain is location, not normalization.}

\begin{table}[t]
\centering \small
\begin{tabular}{llrr}
\toprule
Donor & Reader & Reasoning & Factual \\
\midrule
\multirow{3}{*}{ARC-Challenge} & Motion (raw) & $64.9$ & $46.7$ \\
 & Motion (normalized) & $53.7$ & $36.6$ \\
 & \textbf{Ours} & \textbf{75.6} & \textbf{61.8} \\
\midrule
\multirow{3}{*}{ARC-Easy} & Motion (raw) & $71.7$ & $51.2$ \\
 & Motion (normalized) & $71.6$ & $53.7$ \\
 & \textbf{Ours} & \textbf{77.4} & \textbf{59.5} \\
\bottomrule
\end{tabular}
\caption{Removing magnitude from the motion reader, selection accuracy. Motion (raw) is the
displacement-only baseline; Motion (normalized) reads unit-length displacements, like the location
streams; Ours is the full model. Reasoning is the mean over four reasoning sets, Factual over FACTOR
and VitaminC. Single seed, up to $2{,}000$ items per target.}
\label{tab:normmotion}
\end{table}

\paragraph{Is the gain just the wording?}\label{app:wording}
The state the location streams read holds an answer's wording, so a stream could pass that wording on
in place of location. We repeat both measurements on the streams the detector has learned
(Table~\ref{tab:recover}). The first asks how much of the variation in an answer's wording a
regression can predict from a stream's embedding, where zero would mean the embedding says nothing
about it. The three reach $0.009$ at most. The second asks whether embeddings that sit near one
another belong to answers that share wording, where $1$ would mean they are no more alike than
randomly chosen answers. The three reach $1.32$ at most. On the same two measurements the state a
single-layer probe reads reaches $0.086$ and $2.26$, and pooling that state over the whole answer
raises it to $0.191$ and $3.13$. Between the two
location readings the order is the one the design predicts, the direction stream passing more than
the region code. \textbf{What the streams restore is location, not the wording the raw state holds.}

\begin{table}[t]
\centering \small
\begin{tabular}{lrrr}
\toprule
Representation & Dim & $R^2$ & Retrieval \\
\midrule
Raw state, mean-pooled & 4096 & 0.191 & 3.13 \\
Raw state, last token & 4096 & 0.086 & 2.26 \\
\midrule
Motion & 256 & 0.007 & 1.01 \\
Direction & 32 & 0.009 & 1.32 \\
Region & 32 & $-0.003$ & 1.05 \\
\bottomrule
\end{tabular}
\caption{How much of the correct answer's wording each representation gives back, on the ARC-Challenge
test set ($1{,}172$ items, one correct answer each). $R^2$ is held-out reconstruction of a latent
semantic embedding of the answer text and Retrieval is the nearest-neighbor text-similarity ratio, so
lower is safer in both columns. Dim is the width of the representation read. The raw-state rows are
the residual stream at layer $14$, the depth the probe reads, which is the basis the motivation uses.
One donor (ARC-Challenge), single seed.}
\label{tab:recover}
\end{table}

\section{Parameter Counts}
\label{app:params}
The detector is small, and the location it restores is smaller still. Table~\ref{tab:params} counts the
trainable parameters of each stream on Llama-3.1-8B. The motion stream is the two-layer bidirectional LSTM
over the $4096$-dimensional displacements, at $4.72$M parameters, and the displacement-only baseline of
prior work is exactly this LSTM with a small classification head, $4.76$M in total. The two location
streams we add are far lighter. The direction stream, one shared projection of each layer's state to a
compact coordinate followed by a small aggregator, is $37.5$K parameters. The region stream, whose bulk is
a single projection of each state to $64$ dimensions before the codebook, is $277$K parameters, alongside a
$128$-entry codebook maintained as a moving-average buffer rather than by gradient. \textbf{Together the
two location streams add $0.31$M parameters, $6.7\%$ of the motion LSTM, so restoring the location costs a
small fraction of reading the motion.} The detector as a whole is small against the frozen model it reads.
Llama-3.1-8B has about $8.0$B parameters, so the full detector, at $5.08$M, is $0.06\%$ of the model it
runs on, the same overhead as the displacement-only method it builds on. The counts scale with the model's
residual-stream width.

\begin{table}[t]
\centering \small
\begin{tabular}{lr}
\toprule
Component & Parameters \\
\midrule
Motion stream (bidirectional LSTM) & $4{,}722{,}688$ \\
Direction stream & $37{,}536$ \\
Region stream & $276{,}640$ \\
Classification head & $41{,}217$ \\
\midrule
Full detector & $5{,}078{,}081$ \\
Base model, Llama-3.1-8B (frozen) & ${\sim}8.0$B \\
\bottomrule
\end{tabular}
\caption{Trainable parameter counts on Llama-3.1-8B; the four components sum to the full detector, which
runs on top of the frozen base model. The motion stream is the displacement bi-LSTM of prior work, and the
displacement-only baseline is that LSTM with its own head ($4{,}755{,}713$ parameters). The two location
streams together add $314{,}176$ parameters, $6.7\%$ of the motion LSTM, and the full detector is $0.06\%$
of the base model it reads. The region stream's $128$-entry codebook is a moving-average buffer and is not
counted here.}
\label{tab:params}
\end{table}

\section{Region-Code Usage}
\label{app:regionusage}
\begin{figure*}[t]
\centering
\includegraphics[width=1.99\columnwidth]{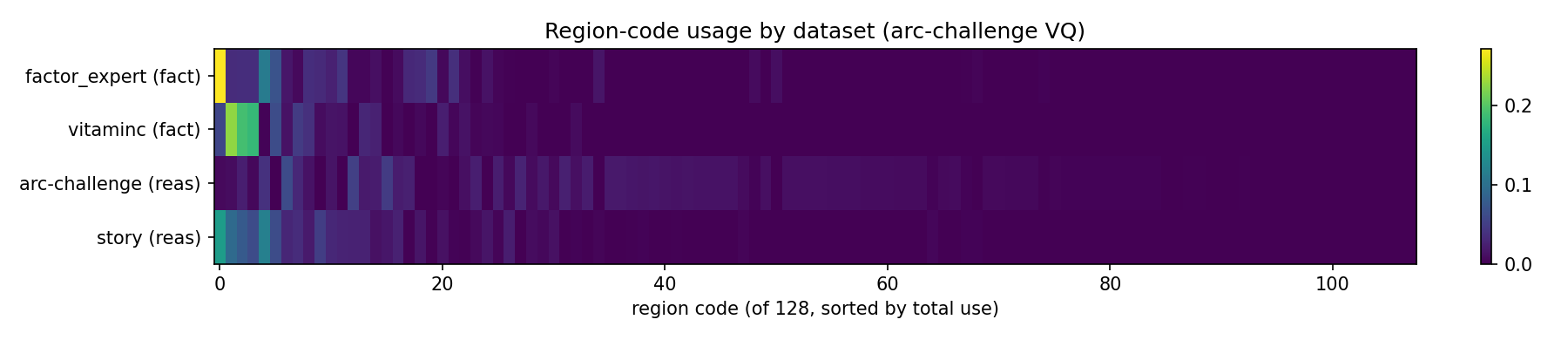}
\caption{Region-code usage by dataset, one row per dataset over the $128$ codebook entries. Factual sets
concentrate on a subset of the codes reasoning sets use. One donor's codebook (ARC-Challenge), up to
$800$ items per dataset.}
\label{fig:region-heatmap}
\end{figure*}

Figure~\ref{fig:region-heatmap} shows which of the $128$ region codes each dataset uses. The factual
rows concentrate in a few bright bands that all lie inside the reasoning rows' support, and no code is
factual-only, which is the usage pattern behind the shared-vocabulary reading in the main text.

Before quantization, a held-out linear read of the raw state reaches AUC $0.92$ on FACTOR-expert and
$0.61$ on FACTOR-wiki (Figure~\ref{fig:region-separation}). Table~\ref{tab:region} shows that part of
this signal survives as a discrete region code.

\begin{table}[t]
\centering \small
\begin{tabular}{lrrr}
\toprule
Dataset & Codes used & Region$\,\to\,$correct & Base rate \\
\midrule
ARC-Challenge & 105 & 81.2 & 75.0 \\
Story Cloze & 42 & 88.2 & 50.0 \\
FACTOR-expert & 41 & 76.1 & 75.0 \\
VitaminC & 32 & 62.8 & 54.4 \\
\bottomrule
\end{tabular}
\caption{Region-code signal: the number of used codes and accuracy from a code-only majority vote.}
\label{tab:region}
\end{table}

\section{Cross-Model Full Results}
\label{app:crossmodel}
Both results in the body come from Llama-3.1-8B. We rerun the pipeline on two more models, a dense
Qwen2.5-14B and a Qwen3-30B-A3B mixture of experts, training the same three readers on the same three
donors, with the layer set read in each model given in Appendix~\ref{app:models}. On reasoning the
displacement reader leads the probe by $2.7$ points on Llama-3.1-8B, the two are level on
Qwen2.5-14B, and on the mixture of experts the displacement reader trails by $2.5$
(Table~\ref{tab:crossmodel}). The full model is
the best of the three readers on reasoning and on factual transfer for all three models. Its margin
over whichever baseline is stronger for a given model is $8$ to $14$ points on reasoning and $11$ to
$16$ on the two factual sets. \textbf{The lead is not inherited from the displacement reader.}

Tables~\ref{tab:crossmodel-reasoning} and~\ref{tab:crossmodel-factual} give the per-donor grids behind the
cross-model summary (Table~\ref{tab:crossmodel}). The full model is the best of the three readers in
every donor cell of both grids, with one
exception on the mixture of experts, where the motion-only reader edges it on ARC-Challenge under the
ARC-Easy donor. Both
models train on the three shared donors and read six evenly spaced layers, single seed, up to $2{,}000$
items per target.

\begin{table}[t]
\centering \small
\setlength{\tabcolsep}{5pt}
\begin{tabular}{llrr}
\toprule
Model & Method & Reasoning & Factual \\
\midrule
\multirow{3}{*}{Llama-3.1-8B} & Linear Probe & 63.7 & 46.2 \\
 & Motion (TaT) & 66.4 & 48.5 \\
 & \textbf{Ours} & \textbf{74.7} & \textbf{59.5} \\
\midrule
\multirow{3}{*}{Qwen2.5-14B} & Linear Probe & 68.8 & 47.2 \\
 & Motion (TaT) & 68.6 & 46.4 \\
 & \textbf{Ours} & \textbf{82.3} & \textbf{63.6} \\
\midrule
\multirow{3}{*}{Qwen3-30B-A3B} & Linear Probe & 74.1 & 49.7 \\
 & Motion (TaT) & 71.6 & 50.3 \\
 & \textbf{Ours} & \textbf{82.2} & \textbf{62.2} \\
\bottomrule
\end{tabular}
\caption{Cross-model summary, selection accuracy, cross-donor means over the three shared donors
(ARC-Challenge, ARC-Easy, CommonsenseQA). Reasoning is the mean over the eight reasoning targets,
Factual over FACTOR-wiki and VitaminC. Single seed; the two Qwen models are scored on up to
$2{,}000$ items per target. Best per model in bold.}
\label{tab:crossmodel}
\end{table}

\begin{table*}[t]
\centering \small
\setlength{\tabcolsep}{3pt}
\begin{tabular}{llccccccccc}
\toprule
Donor & Method & ARC-C & ARC-E & OBQA & CSQA & SIQA & HSwag & MMLU & Story & Mean \\
\midrule
\multicolumn{11}{l}{\textit{Qwen2.5-14B (dense, 48 layers)}} \\
\midrule
    \multirow{3}{*}{ARC-C} & Linear Probe & 82.8 & 90.5 & 78.0 & 59.8 & 66.5 & 71.3 & 55.3 & 91.6 & 74.5 \\
     & Motion (TaT) & 75.7 & 85.3 & 59.6 & 46.3 & 62.3 & 50.9 & 52.6 & 74.9 & 63.4 \\
     & \textbf{Ours} & \textbf{88.3} & \textbf{94.3} & \textbf{86.6} & \textbf{73.7} & \textbf{73.6} & \textbf{84.1} & \textbf{65.0} & \textbf{94.3} & \textbf{82.5} \\
\midrule
    \multirow{3}{*}{ARC-E} & Linear Probe & 80.0 & 91.9 & 80.2 & 64.3 & 61.6 & 64.0 & 56.2 & 87.4 & 73.2 \\
     & Motion (TaT) & 84.6 & 92.6 & 74.0 & 58.6 & 68.4 & 57.2 & 58.3 & 75.3 & 71.1 \\
     & \textbf{Ours} & \textbf{85.4} & \textbf{95.6} & \textbf{86.4} & \textbf{74.4} & \textbf{74.1} & \textbf{84.7} & \textbf{64.6} & \textbf{94.4} & \textbf{82.4} \\
\midrule
    \multirow{3}{*}{CSQA} & Linear Probe & 64.7 & 78.2 & 70.8 & 76.2 & 65.5 & 22.9 & 48.0 & 42.5 & 58.6 \\
     & Motion (TaT) & 74.4 & 85.0 & 80.8 & 82.4 & 72.0 & 47.8 & 53.3 & 73.5 & 71.2 \\
     & \textbf{Ours} & \textbf{85.8} & \textbf{94.0} & \textbf{87.6} & \textbf{84.7} & \textbf{73.4} & \textbf{76.7} & \textbf{63.7} & \textbf{90.6} & \textbf{82.1} \\
\midrule
\multicolumn{11}{l}{\textit{Qwen3-30B-A3B (MoE, 48 layers)}} \\
\midrule
    \multirow{3}{*}{ARC-C} & Linear Probe & 83.4 & 90.6 & 74.8 & 64.5 & 64.9 & 70.4 & 62.3 & 86.4 & 74.7 \\
     & Motion (TaT) & 84.3 & 92.0 & 71.8 & 49.5 & 62.4 & 57.4 & 60.8 & 83.3 & 70.2 \\
     & \textbf{Ours} & \textbf{88.0} & \textbf{95.5} & \textbf{86.8} & \textbf{71.3} & \textbf{71.1} & \textbf{78.7} & \textbf{66.6} & \textbf{95.1} & \textbf{81.6} \\
\midrule
    \multirow{3}{*}{ARC-E} & Linear Probe & 82.9 & 94.1 & 77.6 & 62.3 & 61.9 & 66.8 & 63.8 & 86.4 & 74.5 \\
     & Motion (TaT) & \textbf{87.7} & 94.9 & 79.0 & 66.0 & 66.2 & 66.1 & 63.6 & 92.0 & 76.9 \\
     & \textbf{Ours} & 87.5 & \textbf{96.6} & \textbf{87.8} & \textbf{73.6} & \textbf{69.5} & \textbf{79.7} & \textbf{67.3} & \textbf{94.8} & \textbf{82.1} \\
\midrule
    \multirow{3}{*}{CSQA} & Linear Probe & 81.1 & 92.1 & 81.0 & 79.4 & 65.5 & 38.2 & 61.2 & 85.8 & 73.0 \\
     & Motion (TaT) & 74.9 & 84.3 & 80.8 & 81.2 & 68.6 & 41.8 & 57.9 & 51.6 & 67.6 \\
     & \textbf{Ours} & \textbf{86.3} & \textbf{95.2} & \textbf{88.6} & \textbf{83.5} & \textbf{72.0} & \textbf{78.3} & \textbf{66.3} & \textbf{92.2} & \textbf{82.8} \\
\bottomrule
\end{tabular}
\caption{Cross-model reasoning transfer, selection accuracy $\times 100$, per donor and evaluation target.
Each detector trains on the donor row and is evaluated on every target column. Best per cell in bold; Ours
is best in every cell except one (Qwen3-30B-A3B, ARC-Easy donor, ARC-Challenge target). Single seed, up to
$2{,}000$ items per target.}
\label{tab:crossmodel-reasoning}
\end{table*}

\begin{table*}[t]
\centering
\begin{tabular}{llrrr}
\toprule
Donor & Method & FACTOR-wiki & VitaminC & Mean \\
\midrule
\multicolumn{5}{l}{\textit{Qwen2.5-14B (dense, 48 layers)}} \\
\midrule
    \multirow{3}{*}{ARC-C} & Linear Probe & 37.2 & 46.1 & 41.6 \\
     & Motion (TaT) & 29.1 & 63.2 & 46.2 \\
     & \textbf{Ours} & \textbf{51.1} & \textbf{75.2} & \textbf{63.1} \\
\midrule
    \multirow{3}{*}{ARC-E} & Linear Probe & 37.3 & 68.9 & 53.1 \\
     & Motion (TaT) & 29.8 & 74.0 & 51.9 \\
     & \textbf{Ours} & \textbf{51.5} & \textbf{75.2} & \textbf{63.3} \\
\midrule
    \multirow{3}{*}{CSQA} & Linear Probe & 23.7 & 70.1 & 46.9 \\
     & Motion (TaT) & 26.9 & 55.3 & 41.1 \\
     & \textbf{Ours} & \textbf{49.8} & \textbf{78.5} & \textbf{64.1} \\
\midrule
\multicolumn{5}{l}{\textit{Qwen3-30B-A3B (MoE, 48 layers)}} \\
\midrule
    \multirow{3}{*}{ARC-C} & Linear Probe & 36.9 & 64.1 & 50.5 \\
     & Motion (TaT) & 32.4 & 59.1 & 45.8 \\
     & \textbf{Ours} & \textbf{49.1} & \textbf{73.2} & \textbf{61.1} \\
\midrule
    \multirow{3}{*}{ARC-E} & Linear Probe & 34.1 & 55.6 & 44.9 \\
     & Motion (TaT) & 40.9 & 72.9 & 56.9 \\
     & \textbf{Ours} & \textbf{47.8} & \textbf{74.8} & \textbf{61.3} \\
\midrule
    \multirow{3}{*}{CSQA} & Linear Probe & 28.9 & 78.7 & 53.8 \\
     & Motion (TaT) & 32.4 & 64.4 & 48.4 \\
     & \textbf{Ours} & \textbf{48.0} & \textbf{80.1} & \textbf{64.1} \\
\bottomrule
\end{tabular}
\caption{Cross-model factual transfer, selection accuracy $\times 100$, reasoning-trained detectors
evaluated zero-shot on two length-clean factual sets. Best per cell in bold. Single seed, up to $2{,}000$
items per target.}
\label{tab:crossmodel-factual}
\end{table*}

\section{Where in the Network Does the Location Have to Be Read?}
\label{app:window}
The six depths the two location streams read are taken from the literature on the roles layers play
at different depths of the network, rather than from any measurement on the trained detector. Early
layers group tokens by surface form, late layers work in the space of the token about to be emitted,
and representations come loose from the wording in between. That settles a band and not a set of
layers within it. We retrain the full model five times on the ARC-Challenge donor, changing only the
depths the two location streams read (Table~\ref{tab:window}). Three arms take one third of the stack
each, a fourth reads every layer, and the fifth is our window.

Depth is the largest lever in the method's configuration. A motion-only reader trained the same way
reaches $63.1$ on the reasoning mean, and reading the early third takes it to $63.7$, so the first
third of the network is worth $0.6$ points of location. The middle third is worth $12.5$. Across the
three bands the score rises and then falls, $63.7$ early, $75.6$ middle and $69.5$ late, which is the
shape the three roles predict. Six layers of the middle third also read higher than all $32$, $75.6$
against $74.0$, so which depths are read matters more than how many. Reading all $32$ layers changes
the parameter count as well, so that arm speaks to how many depths are needed and not to which.

Our window is not the best of the five. The middle third reads $3.0$ points above it on the reasoning
mean and is ahead on $11$ of the $12$ targets, and the ordering holds when both arms are retrained.
We keep our window because it was fixed before any of these runs and every number in this paper is
measured with it. \textbf{The gain rests on reading the middle of the network, and not on a window
selected to produce it.} The two axes want different depths. Reading all $32$ layers gives the best
factual mean, $63.7$ against the middle third's $61.3$, and most of that sits on FACTOR-expert, $72.0$
against $64.4$. The factual sets use the late, output-adjacent layers that reasoning does not need,
which is the same place the model's own likelihood reads them from.

\begin{table}[t]
\centering \small
\begin{tabular}{llrr}
\toprule
Window & Depths & Reasoning & Factual \\
\midrule
Early third & $0$--$10$ & 63.7 & 52.5 \\
Middle third & $11$--$21$ & \textbf{75.6} & 61.3 \\
Late third & $22$--$31$ & 69.5 & 57.3 \\
\textbf{Ours} & $8$--$28$ & 72.6 & 60.3 \\
All layers & $0$--$31$ & 74.0 & \textbf{63.7} \\
\bottomrule
\end{tabular}
\caption{The depths the location streams read, selection accuracy $\times 100$, on the ARC-Challenge
donor and the full test sets. The first three arms and ours each read six layers spread across the
band shown and are identical in size; the last reads every layer and is not. Reasoning is the mean
over the eight reasoning targets and Factual over the four factual ones. All five arms share one
training configuration and one evaluation basis and differ only in the depths the two location
streams read. Our window is retrained here under that shared configuration rather than taken from
Table~\ref{tab:main}, so the five arms are directly comparable and the comparison is read within
the sweep. Best per column in bold.}
\label{tab:window}
\end{table}

\section{Detailed Limitations}
\label{app:limitations}
Our detector is only ever scored on selection among the candidates an item supplies. It is never
asked whether a single answer standing on its own is correct, so we cannot say how its score behaves
as an absolute judgment. It needs the model's residual stream, one forward pass per candidate, and a
labeled donor benchmark, and it is trained for the model it reads.

Everything the detector reads comes from the model, so it is bounded by what the model itself
distinguishes. Where the state holds little about whether an answer is correct, no reading of the
trajectory supplies it. A linear read of the raw state tells a correct completion from an incorrect
one at an AUC of $0.92$ on FACTOR-expert and $0.61$ on FACTOR-wiki
(Figure~\ref{fig:region-separation}), and every method we compare, ours included, reads FACTOR-wiki
lower than FACTOR-expert. Reading the model is also what makes the bound measurable. The same read
reports how much a model holds about correctness on a new set before a detector is trained for it.
The streams read activations a forward pass already produces, so the three readings are available
wherever a model scores an answer, and not only in analysis after the fact.

The location streams keep the direction of a state and set its magnitude aside, so what they pass is
an orientation on the unit sphere rather than a position in activation space. Reading the
unnormalized state in place of the direction scored lower in our development runs, so the magnitude
is dropped rather than modeled. Some of the correctness signal may sit there, and a reading that
keeps it as a quantity of its own is untested.

Our fusion does not use everything the three readings hold. A per-item oracle over them stands $3$
to $21$ points of selection accuracy above the trained combination on the four sets we decompose
(Appendix~\ref{app:latefusion}, Table~\ref{tab:complement}), which bounds what a rule choosing between
the streams item by item could add. Closing that gap asks for a better gate over the three streams
rather than another fixed combining rule, since per-item gating and per-step injection score within
$0.005$ AUC of concatenation (Appendix~\ref{app:design}). What such a gate would add beyond accuracy is
a reading of the detector itself. Ours reports which stream is right only after the fact, through a
variant trained to expose one, whereas a gate that weighs the three on each item would state at
prediction time which reading a decision rests on, and so whether the motion of a computation or the
coarse or fine location of its state settles a given case. A detector that reports that is an
instrument for asking which aspect of the model's computation a judgment depends on.

\end{document}